\documentclass[letterpaper]{article} % DO NOT CHANGE THIS
\usepackage[preprint]{aaai2027}  % Public preprint: show authors, omit AAAI footer
\usepackage[hyphens]{url}  % DO NOT CHANGE THIS
\usepackage{graphicx} % DO NOT CHANGE THIS
\usepackage{natbib}  % DO NOT CHANGE THIS AND DO NOT ADD ANY OPTIONS TO IT
\usepackage{caption} % DO NOT CHANGE THIS AND DO NOT ADD ANY OPTIONS TO IT
\usepackage{algorithm}
\usepackage{algorithmic}
\usepackage{newfloat}
\usepackage{listings}
\DeclareCaptionStyle{ruled}{labelfont=normalfont,labelsep=colon,strut=off} % DO NOT CHANGE THIS
\lstdefinestyle{finalprompt}{%
  basicstyle={\fontsize{11}{13}\selectfont\ttfamily},
  numbers=none,
  xleftmargin=0pt,
  framexleftmargin=3pt,
  framexrightmargin=3pt,
  aboveskip=0pt,
  belowskip=0pt,
  frame=single,
  showstringspaces=false,
  columns=fullflexible,
  keepspaces=true,
  tabsize=2,
  breaklines=true,
  breakatwhitespace=true}
\floatstyle{ruled}
\newfloat{listing}{tb}{lst}{}
\floatname{listing}{Listing}
\usepackage{booktabs}
\usepackage{array}
\usepackage{amsmath}
\usepackage{amssymb}
\usepackage{amsthm}
\usepackage{placeins}

\theoremstyle{definition}
\newtheorem{definition}{Definition}

\usepackage{xcolor}
\usepackage{colortbl}
\usepackage{tikz}

\definecolor{scorbluegray}{HTML}{DCE4E7}
\definecolor{scorred}{HTML}{D9828F}
\definecolor{oursrow}{HTML}{EEF5FC}

\definecolor{circleTealFill}{HTML}{CCE6E6}
\definecolor{circleTealLine}{HTML}{008080}
\definecolor{circlePinkFill}{HTML}{F2CDD9}
\definecolor{circlePinkLine}{HTML}{BF0140}
\definecolor{circlePurpleFill}{HTML}{E7DFFF}
\definecolor{circlePurpleLine}{HTML}{6546C4}

\definecolor{circleBlueFill}{HTML}{CCCCFF}
\definecolor{circleBlueLine}{HTML}{0000FF}
\definecolor{circleOrangeFill}{HTML}{FFE6CC}
\definecolor{circleOrangeLine}{HTML}{FF8000}

\newcommand{\stageone}{%
  \tikz[baseline=(n.base)]{
    \node[
      circle,
      draw=circleTealLine,
      fill=circleTealFill,
      line width=0.45pt,
      minimum size=1.22em,
      inner sep=0pt,
      outer sep=0pt,
      font=\small\bfseries
    ] (n) {1};
  }\nobreak\hspace{0.12em}%
}

\newcommand{\stagetwo}{%
  \tikz[baseline=(n.base)]{
    \node[
      circle,
      draw=circlePinkLine,
      fill=circlePinkFill,
      line width=0.45pt,
      minimum size=1.22em,
      inner sep=0pt,
      outer sep=0pt,
      font=\small\bfseries
    ] (n) {2};
  }\nobreak\hspace{0.12em}%
}

\newcommand{\stagethree}{%
  \tikz[baseline=(n.base)]{
    \node[
      circle,
      draw=circlePurpleLine,
      fill=circlePurpleFill,
      line width=0.45pt,
      minimum size=1.22em,
      inner sep=0pt,
      outer sep=0pt,
      font=\small\bfseries
    ] (n) {3};
  }\nobreak\hspace{0.12em}%
}

\title{SCoR: A Hierarchical Framework for\\Forecasting Relations Between Scientific Concepts}
\author{
    Jingze Wang\equalcontrib,
    Fred Sun\equalcontrib,
    Shangqi Guo\corresponding
}
\affiliations{
    Center for Brain-Inspired Computing Research, Tsinghua University
}

\newcommand{\bestscore}[1]{\ensuremath{\pmb{#1}}}
\newcommand{\secondscore}[1]{\ensuremath{\underline{#1}}}
\newcommand{\bestsciscore}[2]{\ensuremath{\pmb{#1{\times}10^{#2}}}}
\newcommand{\secondsciscore}[2]{\ensuremath{\underline{#1{\times}10^{#2}}}}
\newcommand{\bestlift}[1]{\ensuremath{\pmb{#1{\times}}}}
\newcommand{\secondlift}[1]{\ensuremath{\underline{#1{\times}}}}

\begin{document}
\maketitle

\begin{abstract}
Anticipating emerging research directions is a critical goal of AI-assisted
science, helping researchers identify promising questions and formulate
testable hypotheses. Existing methods mainly predict which concepts will
co-occur in future papers. Yet co-occurrence captures shared attention rather
than the scientific meaning of a connection, which lies in the relation
expressed between the concepts---for example, whether one method \textit{uses},
\textit{combines}, \textit{replaces}, or \textit{contradicts} another.
Forecasting such relations
can provide a more precise and interpretable view of how science evolves. We
formulate research-direction discovery as hierarchical scientific-relation
forecasting over a shared candidate-pair space, comprising three temporally
aligned tasks: first co-occurrence, first scientific-relation formation, and
relation type at formation. To instantiate this formulation, we construct
\textbf{SCoR-Graph} from 187{,}848 \texttt{cs.CV} papers published between
2017 and 2026, yielding 270{,}687 consolidated concepts, 7.45 million
co-occurrence edges, and 615{,}036 typed, directed relation edges. From
cutoff-specific graph snapshots, we derive \textbf{SCoR-Bench}, a
leakage-audited benchmark for these three forecasting capabilities, with
expert-verified gold labels for the entire relation-type test set. We further
introduce \textbf{HiSCoR}, a task-adapted model family that models relation
emergence as a temporally evolving and hierarchically constrained process by
encoding pre-cutoff event histories and conditioning relation formation on
future co-occurrence. On the held-out 2025--2026 window, HiSCoR achieves an
AUROC of 0.9515, representing a 2.4\% relative improvement over the strongest
temporal-graph baseline, while improving population-AUPRC by 14.0\%; its
task-adapted relation-type variant achieves a
Macro-AUROC of 0.7795.
Ablations show that semantic, co-occurrence, and typed-relation views provide
complementary predictive evidence. SCoR advances research-direction
forecasting from predicting which concepts will co-occur to anticipating
whether and how evidence-backed scientific relations will emerge.
\end{abstract}

% =====================================================================
\section{Introduction}

Scientific advances often emerge when previously separate concepts, methods,
and findings become connected through meaningful relations: a neural
architecture may be adapted to a new task, techniques from different
subfields may be combined, or new evidence may challenge an established
approach. Anticipating such connections before they appear in the literature
could help AI systems surface emerging research directions and propose
testable hypotheses ahead of the field. A productive line of research
operationalizes this goal through a concept co-occurrence graph: concepts are
extracted from historical papers, linked whenever they appear together, and
previously unconnected pairs are ranked by how likely they are to co-occur in
a future window
\citep{krenn2020predicting,krenn2023forecasting,marwitz2026predicting}.
Because each prediction is checked against papers published after a fixed
cutoff, research-direction discovery becomes a temporally verifiable
forecasting task.

Co-occurrence, however, records only that two concepts drew joint attention.
Two concepts may share a paper because one method uses another, two
techniques are combined, one approach replaces another, or a new result
contradicts an earlier claim; these relations differ in meaning, direction,
and value, yet an untyped graph collapses them into the same edge.
Forecasting whether and how such a relation forms therefore moves
research-direction discovery beyond topic-level attention and toward the
development of scientific knowledge.

Making this shift raises three requirements. First, scientific concepts form
an open, evolving vocabulary: a vision transformer may appear as ``ViT,''
``vision transformer,'' or a named variant. Mentions must therefore be
reconciled into identities that remain stable across temporal snapshots. Second,
relations live in the full text and must be grounded in textual evidence
establishing whether a relation exists, its type, its direction, and when it
first formed. Third, the prediction targets occupy nested sample spaces: a
relation can occur only within a co-occurrence event, and relation type is
defined only for relation-positive pairs, so treating events outside these
conditioning spaces as negatives introduces sample-selection bias.

Existing paradigms leave this problem open: concept-graph forecasting
predicts only untyped co-occurrence
\citep{krenn2023forecasting,marwitz2026predicting}, literature-based
discovery does not treat first relation emergence as a temporally held-out
forecasting event \citep{swanson1986undiscovered,sybrandt2020agatha}, and
temporal knowledge-graph forecasting assumes a fixed vocabulary and scores
triples independently \citep{trivedi2017knowevolve,li2021regcn}. What is
missing is a temporally grounded formulation that makes the event hierarchy
and its task-specific sample spaces explicit.

\begin{figure}[!t]
    \centering
    \includegraphics[width=0.85\columnwidth]{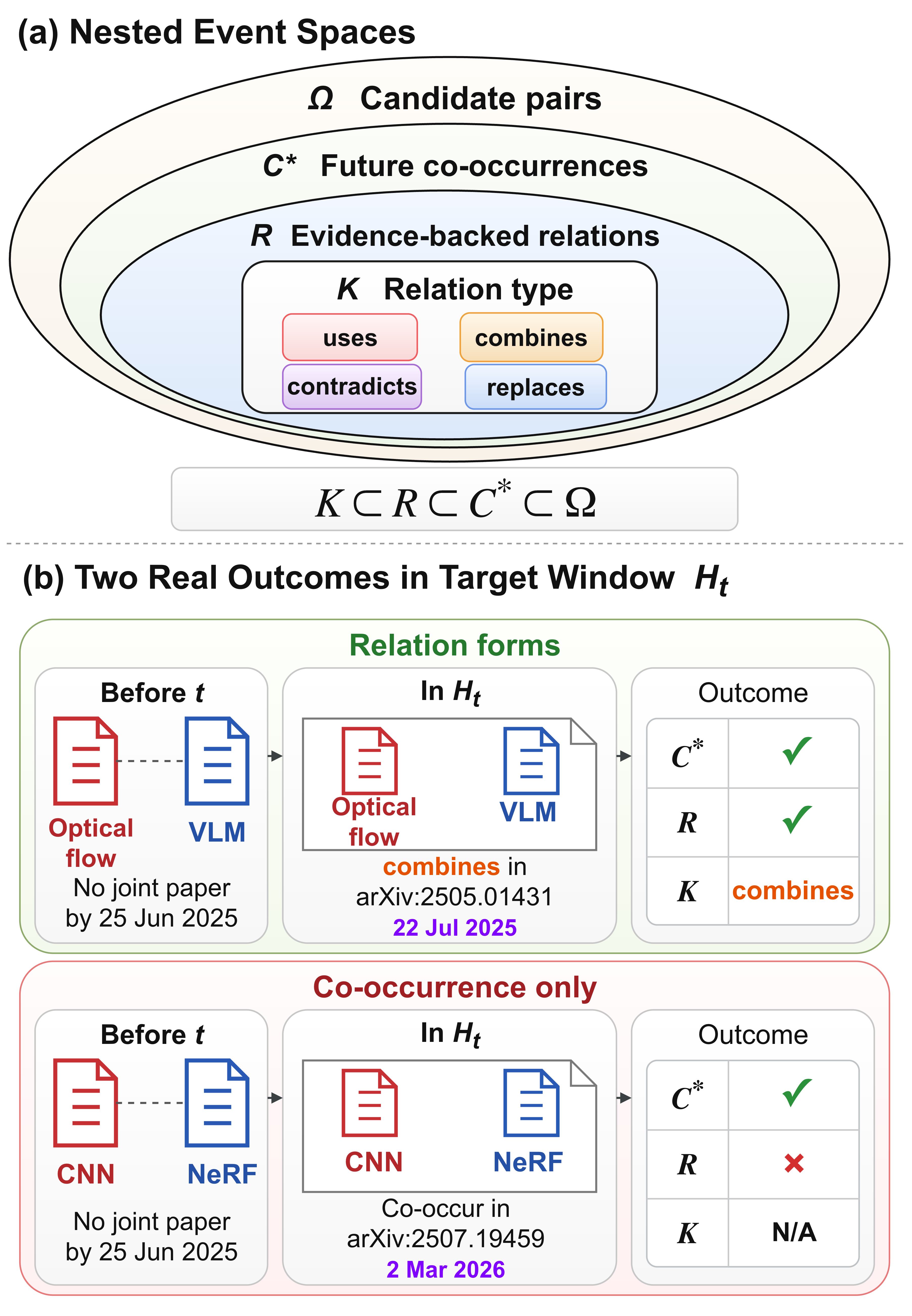}
    \caption{Nested event spaces for scientific-relation forecasting. (a)
    Within the candidate space $\Omega$, pairs that co-occur in the target
    window $\mathcal{H}_t$ form $C^\star$; relation-positive pairs form the
    nested subset $\mathcal{R}$, on which alone relation types are defined.
    $T_0$--$T_2$ evaluate first co-occurrence, first relation formation, and
    relation type. (b) Both example pairs co-occur, yet only one forms an
    evidence-backed relation: optical flow is explicitly combined with a
    VLM in their joint paper (\textsc{combines}), whereas CNN and NeRF
    merely appear in the same paper, so no relation type is defined.}
    \label{fig:nested_sample_space}
\end{figure}

To address this gap, we view a research direction not as an isolated future
link, but as a connection that acquires scientific meaning through three
conditionally dependent stages: \stageone\textbf{Target-window
co-occurrence ($C^\star$):} two concepts receive joint attention by
appearing in the same paper; \stagetwo\textbf{Scientific-relation formation
($R$):} their co-occurrence supports a substantive, evidence-backed
relation; and \stagethree\textbf{Relation typing ($K$):} the relation takes
a specific semantic form, such as using, combining with, replacing, or
contradicting another concept. This hierarchy describes conditional
observability rather than a fixed temporal sequence, since all three events
may first become observable in the same paper. The stages carve the
candidate pairs $\Omega$ into the nested event spaces
$K\subset R\subset C^\star\subset\Omega$ of
Figure~\ref{fig:nested_sample_space}(a), and
Figure~\ref{fig:nested_sample_space}(b) shows why the distinction matters:
optical flow and a VLM co-occur and form an evidence-backed
\textsc{combines} relation, whereas CNN and NeRF merely co-occur without a
scientific relation. We operationalize the hierarchy
through temporally aligned forecasts, modeling the first two stages jointly
to preserve their prerequisite relationship while conditioning type
prediction on relation formation, keeping each task inside its valid
conditioning space.

To make the formulation measurable, we construct \textsc{SCoR-Graph}, which
consolidates the concepts of 187{,}848 \texttt{cs.CV} papers (2017--2026)
into semantic, co-occurrence, and typed-relation views, and derive
\textsc{SCoR-Bench}, which turns its cutoff-specific snapshots into
leakage-audited forecasting tasks for the three stages, with rolling temporal
evaluation and expert-verified gold relation-type test labels.

On this benchmark we introduce \textsc{HiSCoR}, a family of task-adapted
forecasters combining two complementary views of relation emergence: the
static scientific state captured by semantic, co-occurrence, and
typed-relation structure, and the dynamics captured by pre-cutoff
concept-pair events. For relation formation, a hierarchy-consistent static
model is fused with a temporal event encoder through constrained logit
residuals, letting dynamic evidence sharpen ranking while preserving
consistency between future co-occurrence and relation formation.

On the held-out 2025--2026 window, HiSCoR attains 0.9515 AUROC and
$3.859\times10^{-3}$ population-AUPRC for relation formation, improving over
the strongest temporal-graph baseline by 2.4\% and 14.0\%, respectively,
with zero hierarchy violations; task-adapted variants likewise lead
AUROC/Macro-AUROC on first co-occurrence and relation type. Matched
diagnostics on the precursor HiSCoR-EF/EF-Type and HiSCoR-State variants
support complementary semantic,
co-occurrence, relation, and temporal-event evidence, but do not decompose
the final HiSCoR gain component by component.

% =====================================================================
\section{Related Work}
\label{sec:related}

\paragraph{Scientific concept discovery and relation extraction.}
Science-of-science and concept-graph methods model how ideas evolve through
temporally held-out link prediction \citep{fortunato2018science}. SemNet,
Science4Cast, and the materials-science model of
\citet{marwitz2026predicting} forecast concept co-occurrences using semantic
and structural evidence
\citep{krenn2020predicting,krenn2023forecasting,marwitz2026predicting};
literature-based discovery similarly ranks latent associations
\citep{swanson1986undiscovered,sybrandt2020agatha}. In parallel, scientific
information extraction and scholarly knowledge graphs recover relations
already stated in observed documents
\citep{luan2018multitask,jain2020scirex,zhang2024scier,jaradeh2019orkg}.
These lines either predict untyped links or encode observed relations. SCoR
instead forecasts when an evidence-backed relation first forms and which
coarse type it takes.

\paragraph{Dynamic graph and temporal knowledge-graph forecasting.}
GraphMixer and DyGFormer encode pre-cutoff interaction histories for future
link prediction \citep{cong2023graphmixer,yu2023dygformer}, whereas temporal
knowledge-graph models forecast timestamped typed facts from relational
histories \citep{trivedi2017knowevolve,li2021regcn,zhu2021cygnet,
liu2022tlogic,park2022evokg,dong2023daemon,gastinger2024recb}.
Cross-graph methods such as ULTRA further transfer relational reasoning
between knowledge graphs \citep{galkin2024ultra}. These methods typically
assume fixed entity and relation vocabularies and score future links or facts
independently. Our setting instead consolidates an evolving concept
vocabulary and couples co-occurrence, relation formation, and relation type
in one temporal hierarchy.

\paragraph{Entire-space and hierarchical multi-task modeling.}
In post-click conversion modeling, conversion is observable only after a
click. ESMM handles this selection pattern by factorizing an entire-space
probability into an exposure event and a conditional event \citep{ma2018entire}.
HiSCoR transfers this principle to the event hierarchy of
\S\ref{sec:formulation}, and unlike standard ESMM combines it with static
semantic, co-occurrence, and typed-relation evidence plus pre-cutoff event
dynamics.

% =====================================================================
\section{Method}
\label{sec:method}

\begin{figure*}[!t]
    \centering
    \includegraphics[width=0.87\textwidth]{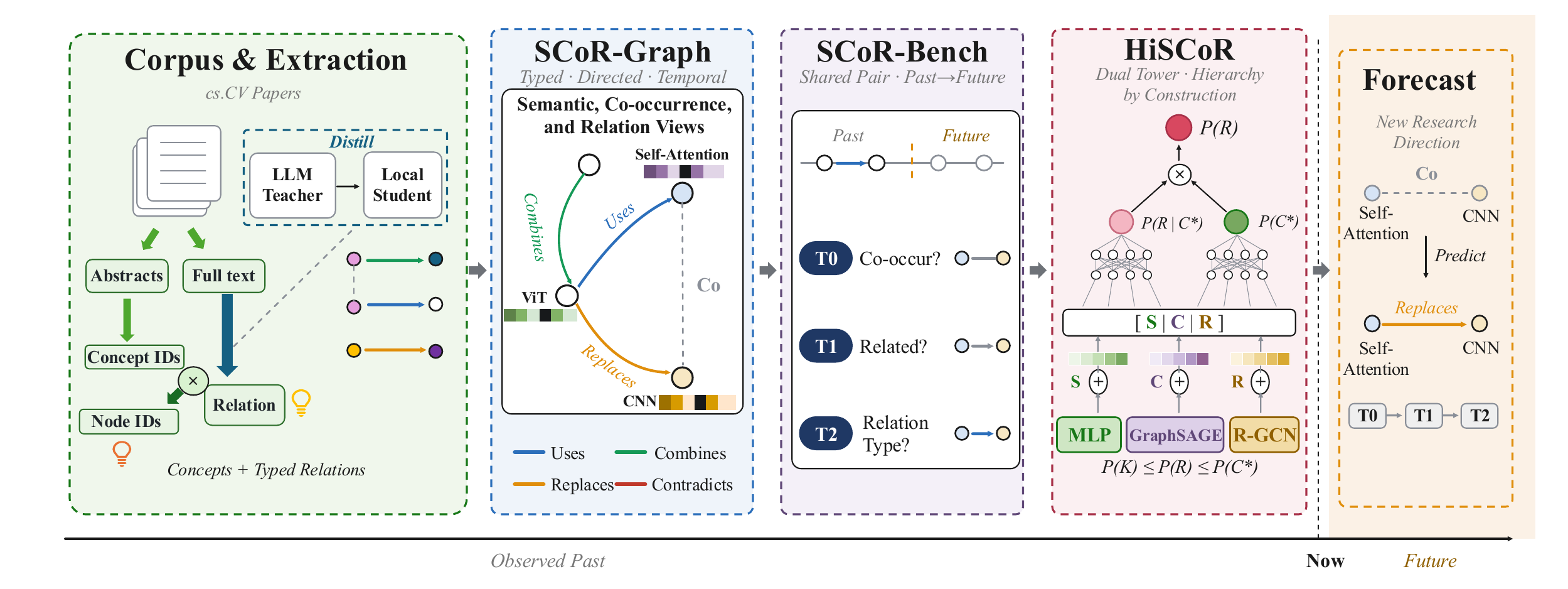}
    \caption{End-to-end overview of SCoR. Literature is converted into
    \textsc{SCoR-Graph}, whose semantic, co-occurrence, and typed-relation
    views support the \textsc{SCoR-Bench} tasks; \textsc{HiSCoR} couples
    target-window co-occurrence with relation formation through a
    hierarchy-constrained dual tower, with relation type evaluated
    conditional on formation.}
    \label{fig:scor_pipeline}
\end{figure*}

SCoR is organized around a single question: given only the literature
available at a cutoff, will two concepts form an evidence-backed relation
within the target window, and of what type?
Figure~\ref{fig:scor_pipeline} summarizes how the design meets the three
requirements identified in the introduction. \S\ref{sec:formulation}
formalizes discovery as three temporally aligned tasks over explicit
conditioning populations, respecting the nested sample spaces;
\S\ref{sec:scor_graph} consolidates the literature into SCoR-Graph,
whose semantic, co-occurrence, and typed-relation views supply
cutoff-safe evidence; \S\ref{sec:scor_bench} turns its cutoff-specific
snapshots into the leakage-audited SCoR-Bench with
population-preserving evaluation; and \S\ref{sec:hiscor} presents
HiSCoR, which fuses the static graph state with pre-cutoff event
dynamics under the event hierarchy.

% ===== inlined: sections/method_revised.tex =====
\subsection{Hierarchical Scientific-Relation Forecasting}
\label{sec:formulation}

Forecasting a scientific relation is not equivalent to forecasting an
untyped link: it must respect the nested event hierarchy of
Fig.~\ref{fig:nested_sample_space}. A different issue arises when
constructing a \emph{first-event} benchmark: pairs that have already
experienced the target event before the cutoff must be excluded rather than
relabeled as negatives. We keep these two ideas separate throughout.

\begin{definition}[Cutoff-specific first-event forecast]
Let \(t\) be a cutoff, \(\mathcal H_t=(t,t+\Delta]\) the forecast
horizon, and \(\mathcal U_t\) the unordered pairs of concepts observed by
\(t\). Let \(\tau_0=\tau_C\) and \(\tau_1=\tau_R\) denote the first
co-occurrence and first trusted-relation times. For
\(j\in\{0,1\}\), the eligible population and its label are
\begin{equation}
\begin{aligned}
\Omega_j(t)
&=\{\{u,v\}\in\mathcal U_t:\tau_j(u,v)>t\},\\
y^{(j)}_{uv,t}
&=\mathbf 1[t<\tau_j(u,v)\le t+\Delta].
\end{aligned}
\label{eq:forecasting_protocol}
\end{equation}
T2 is defined only for T1-positive pairs; its target is the unambiguous
category on the pair's first trusted-relation day.
\end{definition}

Equation~\ref{eq:forecasting_protocol} defines T0 as first
co-occurrence and T1 as first scientific-relation formation. Their
eligibility sets are not nested: a pair may have co-occurred before
\(t\) yet remain eligible for its first scientific relation. The nested
structure instead concerns events \emph{inside} \(\mathcal H_t\).
Specifically, let \(C^\star\) denote any target-window co-occurrence,
\(R\) a target-window relation formation, and \(K\) its type. Here
\(C^\star\) differs deliberately from T0 because it need not be the
pair's first co-occurrence. Relation candidates are restricted to
concept pairs extracted from the same paper's abstract, while the full
text is used only to verify and type their relation. Hence every accepted
relation event has a corresponding same-paper co-occurrence event, so
\(R\subseteq C^\star\) holds by construction; an audit over all benchmark
windows found zero violations. This gives
\begin{equation}
\begin{aligned}
P(R{=}1,K{=}k\mid\mathcal{G}_t)
={}&P(C^\star{=}1\mid\mathcal{G}_t)\\
&\times P(R{=}1\mid C^\star{=}1,\mathcal{G}_t)\\
&\times P(K{=}k\mid R{=}1,\mathcal{G}_t).
\end{aligned}
\label{eq:hierarchical_factorization}
\end{equation}
This factorization is the conceptual distinction from three independent
classifiers: it encodes the scientific prerequisite between future events
while leaving the eligibility sets of Eq.~\ref{eq:forecasting_protocol}
unnested. All conditioning evidence in \(\mathcal G_t\) is restricted to
information available by the cutoff.

\subsection{SCoR-Graph}
\label{sec:scor_graph}

\begin{definition}[SCoR-Graph snapshot]
At cutoff \(t\), a SCoR-Graph snapshot is the tuple
\(\mathcal G_t=(\mathcal V_t,\mathbf X_t^{\mathcal S},
\mathcal E_t^{\mathcal C},\mathcal E_t^{\mathcal R})\).
The semantic field \(\mathbf X_t^{\mathcal S}\) stores context and
identity representations. The co-occurrence field
\(\mathcal E_t^{\mathcal C}\) contains undirected, weighted, time-stamped
edges. The scientific-relation field \(\mathcal E_t^{\mathcal R}\)
contains evidence-backed, weighted, time-stamped
\textsc{uses}, \textsc{combines}, \textsc{replaces}, and
\textsc{contradicts} edges, preserving direction for asymmetric
relations.
\end{definition}

The three fields capture complementary signals: concept meaning and usage,
topical proximity and activity, and the scientific roles connecting
concepts. Keeping them distinct allows their predictive contributions to be
measured rather than collapsed into one untyped edge.

\paragraph{From literature to stable concept identities.}
We instantiate SCoR-Graph from 187{,}848 \texttt{cs.CV} papers published
between January 2017 and June 2026. Concept mentions are extracted from
abstracts and conservatively consolidated into stable identities using a
retrieval-and-verification model over surface forms. Each identity stores
two semantic views: a context vector aggregated from the concept's
abstract occurrences, and an identity vector learned to retrieve
equivalent surface forms. Aliases become active only after their first
observed date, so later terminology cannot alter an earlier snapshot.

\paragraph{Co-occurrence and relation evidence.}
Every paper contributes a co-occurrence event for each pair of distinct
concept identities in that paper; repeated support accumulates as an edge
weight. Typed relations are instead extracted from full-text evidence.
An audited generator--discriminator process creates high-precision
supervision, which is distilled into local relation extractors for
corpus-scale processing. Uncertain, negated, comparative, and rare-class
cases are routed back for adjudication. Each accepted instance retains its
source paper, evidence span, timestamp, confidence, type, direction, and
provenance. Thus the large relation layer is auditable distilled
supervision rather than an unqualified human-gold corpus. On a
10{,}000-instance expert-verified audit set, accepted relations achieved
93.4\% relation validity, 90.7\% type correctness, and 89.9\% direction
correctness, supporting the precision of the resulting relation layer.

\paragraph{Temporal graph assembly.}
Concept identities are consolidated before edges are assembled, avoiding
alias-induced self-loops and inflated weights. Co-occurrence support is
aggregated over papers, while relation evidence is deduplicated within
and across papers. Directed relations are expanded into forward and
inverse message channels for graph encoding. Every snapshot truncates
nodes, semantic observations, edge events, and aliases at the same cutoff
\(t\). Through June 2026, the resulting graph contains 270{,}687 concept
nodes, 7.45 million unique co-occurrence edges, and 615{,}036 unique typed
relation edges: 440{,}689 \textsc{uses}, 168{,}982 \textsc{combines},
4{,}487 \textsc{replaces}, and 878 \textsc{contradicts}.

\subsection{SCoR-Bench}
\label{sec:scor_bench}

SCoR-Bench turns SCoR-Graph into a forecasting benchmark with a simple
contract: a model sees only the graph as it existed at a cutoff, and it is
scored on which concept pairs actually connect during the following year,
matching the first-event forecasts of Definition~1. The subtlety is that
these events are extremely rare, so seemingly minor choices, such as how
negative pairs are sampled, can quietly change what is being measured.
SCoR-Bench therefore keeps three ingredients separate: the event to be
predicted, the population of pairs over which it is defined, and the
retrieval procedure that proposes candidate pairs; the following
paragraphs address each in turn.

\paragraph{Labels and rolling horizons.}
All tasks use a fixed one-year horizon. A T0 or T1 negative means only
that the event does not occur within \(\mathcal{H}_t\), not that the pair
will never connect; later events therefore remain negatives in the primary
fixed-window benchmark. T2 is evaluated on relation-positive pairs and
uses three statistically viable categories: \textsc{uses},
\textsc{combines}, and the merged \textsc{replaces}/\textsc{contradicts}
class. Pairs with conflicting first-day categories are excluded, and
endpoint orientation is not a T2 target. We train on the 25 June
2022--2023 and 2023--2024 windows, select models on 2024--2025, and
reserve 2025--2026 for the final test.

\paragraph{Population-preserving evaluation.}
For T0 and T1, the benchmark retains a census of observed positives and
a probability sample of eligible negatives. Every sampled row records
its inclusion probability. We therefore evaluate with inverse-probability
weights, recovering estimates for the eligible population rather than
for an artificially balanced table. The exact estimator is given in
Appendix~\ref{app:population}.
Positive or hard-negative enrichment is permitted only inside the
training DataLoader. It changes gradient exposure, not labels,
validation/test rows, or population weights.

\paragraph{Future-blind retrieval and diagnostic views.}
An operational candidate pool may use only \(\mathcal{G}_t\), including
historical graph proximity, semantic neighbors, and uniformly sampled
active pairs. Its candidate recall is reported separately because it is
an upper bound on end-to-end discovery. A matched controlled view is
retained only for architecture diagnosis, while a retrospective strict
view that removes known delayed events is used only for sensitivity
analysis. Neither controlled nor oracle-conditioned results are
presented as full-space discovery performance.

\paragraph{Trusted relations and verified evaluation.}
The first trusted evidence timestamp defines relation formation.
Trust can arise from teacher adjudication, calibrated agreement between
independent extractors, or high-confidence evidence with retained
provenance; ambiguous instances are not converted into negatives.
Training uses this audited distilled layer. Relation-type test labels are
additionally reviewed against full-text evidence, yielding expert-verified
gold labels for the entire T2 test set.

\subsection{HiSCoR}
\label{sec:hiscor}

\begin{figure}[!h]
    \centering
    \includegraphics[width=\columnwidth]{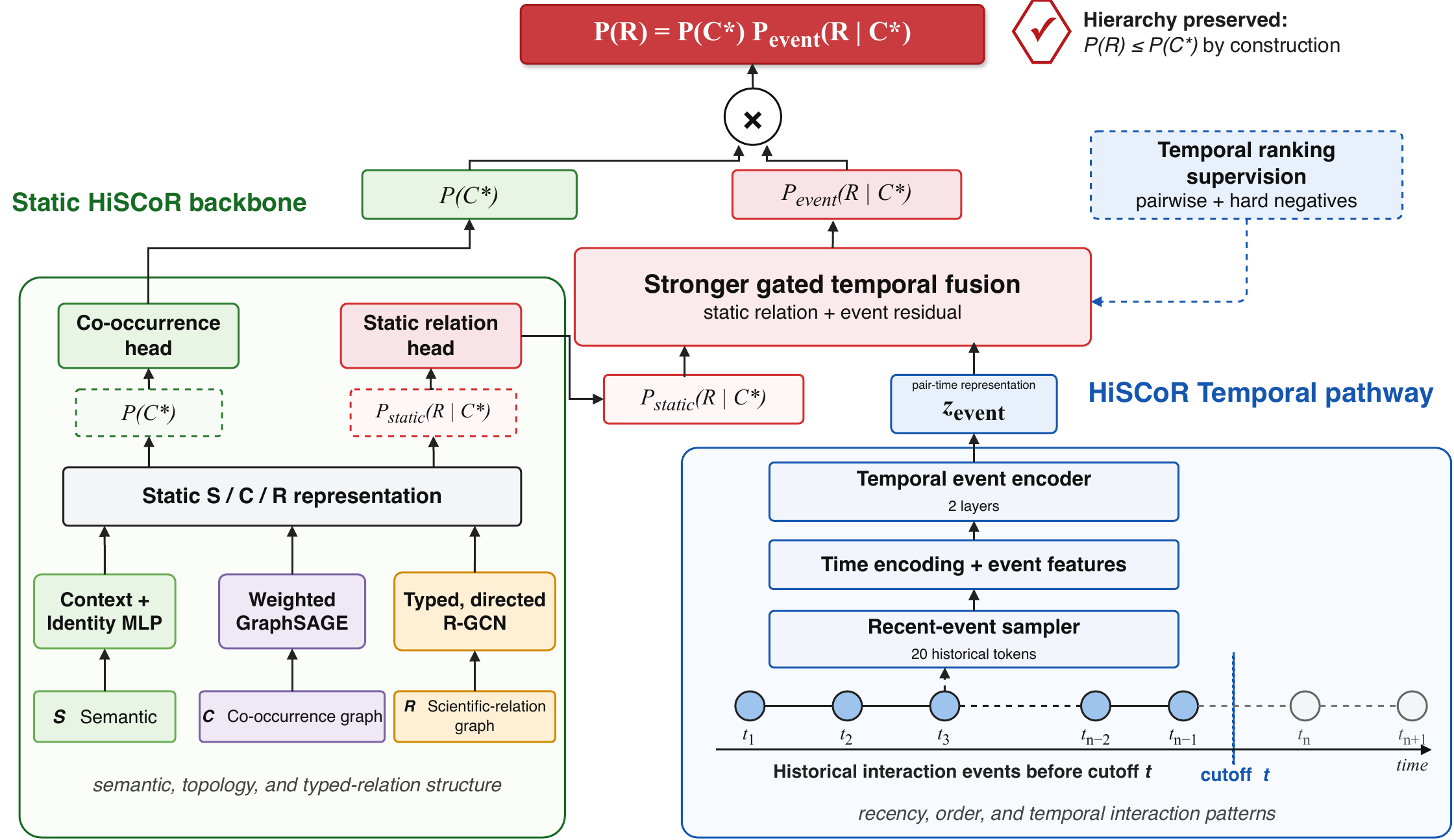}
    \caption{\textbf{HiSCoR for relation formation.}
    A static S/C/R backbone estimates \(P(C^\star)\) and
    \(P_{\mathrm{static}}(R\mid C^\star)\); the event pathway adds a
    gated residual with training-only ranking supervision. Their product
    preserves \(P(R)\leq P(C^\star)\) by construction.}
    \label{fig:hiscor_event}
\end{figure}

HiSCoR is a family of task-adapted predictors sharing the three S/C/R
evidence fields: the static branch asks whether two concepts are
semantically and structurally compatible, while the event branch asks
whether their recent interactions make a new connection imminent. T0
pairs the static state with co-occurrence events, T1 additionally uses
typed-relation events under the hierarchy of
Eq.~\ref{eq:hierarchical_factorization}, and T2 applies a type-sensitive
relation state with a class-wise temporal correction.
Figure~\ref{fig:hiscor_event} summarizes the relation-formation variant.

\paragraph{Hierarchy-constrained parameterization.}
For relation formation, a three-state head distinguishes no
target-window co-occurrence, co-occurrence without relation, and relation
formation, yielding \(p^C=P(C^\star=1\mid\mathcal G_t)\) and a static
conditional estimate \(q^{\mathrm{static}}=P(R=1\mid
C^\star=1,\mathcal G_t)\); unlike two unrelated binary heads, this makes
their shared population explicit. A temporal encoder reads only events
before \(t\) and contributes a gated signed correction to the
\emph{conditional} relation logit:
\begin{equation}
\begin{aligned}
q^{\mathrm{event}}
&=\sigma\!\left(\operatorname{logit}q^{\mathrm{static}}
+\beta\,g(\mathbf z,\mathbf e)\,r(\mathbf z,\mathbf e)\right),\\
p^R&=p^Cq^{\mathrm{event}}\leq p^C .
\end{aligned}
\label{eq:hiscor_core}
\end{equation}
Here \(\mathbf z\) is the static pair state, \(\mathbf e\) the ordered
event state, \(g\in[0,1]\) gates how reliable the event evidence is, and
\(r\in\mathbb R\) supplies its signed adjustment. Because the residual
acts inside \(P(R\mid C^\star)\) rather than on \(P(R)\), dynamics can
refine the relation decision while \(p^R\leq p^C\) holds for every
parameter value: a structural guarantee, not a penalty or
post-processing rule. All encoders, gates, and heads are optimized
jointly. Final T1/T2 HiSCoR variants add a nonzero gate floor and
task-matched hard-negative ranking; we call the original gated variants
HiSCoR-EF and HiSCoR-EF-Type. Inference remains
Eq.~\ref{eq:hiscor_core}; exact settings appear in
Appendix~\ref{app:temporal_refinement}.

\paragraph{S/C/R pair state and event state.}
Each field has a matched encoder: gated fusion of cutoff-safe context and
identity vectors for semantics (how a concept is used, and which surface
forms denote it), a support-weighted GraphSAGE over the co-occurrence
graph for research activity, and a weight-aware relational GCN whose
forward and inverse channels preserve relation type and direction.
Symmetric pair operators combine the endpoint states and concatenate the
normalized fields into \(\mathbf z\), keeping the sources separable for
ablation (Appendix~\ref{app:fields}). The event state \(\mathbf e\)
mixes the pair's recent pre-cutoff events, carrying continuous time and
event attributes, through an MLP-Mixer. T1 consumes both co-occurrence
and typed-relation histories because either can signal an approaching
relation; T0 uses co-occurrence history alone, since relation events do
not define its target, and corrects the first-co-occurrence logit
directly.

\paragraph{Conditional relation-type head.}
T2 estimates \(P(K\mid R,\mathcal G_t)\) and therefore needs a
category-sensitive rather than formation-sensitive relation state:
endpoint-conditioned, EPGNN-style neighborhood aggregation retains
relation type, direction, and support weight; semantic evidence forms
the second principal field; and co-occurrence enters only as a weak
residual. A class-wise temporal residual updates the three category
logits, since recency can support one relation type while suppressing
another. Final HiSCoR-Type adds a nonzero gate floor and balanced
one-vs-rest hard-negative supervision to this residual. Because every
T2 example already satisfies \(R=1\) and endpoint
orientation is not a T2 target (\S\ref{sec:scor_bench}), the model adds
no relation-existence head and treats the graph's directed history as
input evidence rather than a prediction label
(Appendix~\ref{app:type}).

\paragraph{Training objective.}
Each variant optimizes a target matched to its sample space:
population-weighted binary cross-entropy (T0), population-weighted
cross-entropy over the three coherent states (T1), and class-balanced
cross-entropy (T2). Training-only ranking terms increase exposure to
rare positives and hard negatives without altering the evaluation
population (\S\ref{sec:scor_bench}); the complete objectives are listed
in Appendix~\ref{app:objectives}, and optimization and sampling schedules
accompany the experimental protocol.

% =====================================================================
% ===== inlined: sections/experiments_v2.tex =====
% =====================================================================
% experiments_v2.tex: Experiments section (reconstructed from the
% final 7-page PDF; numbers follow the 2026-07-29 unified three-seed
% records together with the SCoR-Bench v2 evidence audit. Test results
% use the natural-population protocol, seeds 17/23/42, and the fixed
% 2025-06-25 -- 2026-06-25 test window).
% =====================================================================

\begin{figure*}[!t]
\centering
\includegraphics[width=0.8675\textwidth]{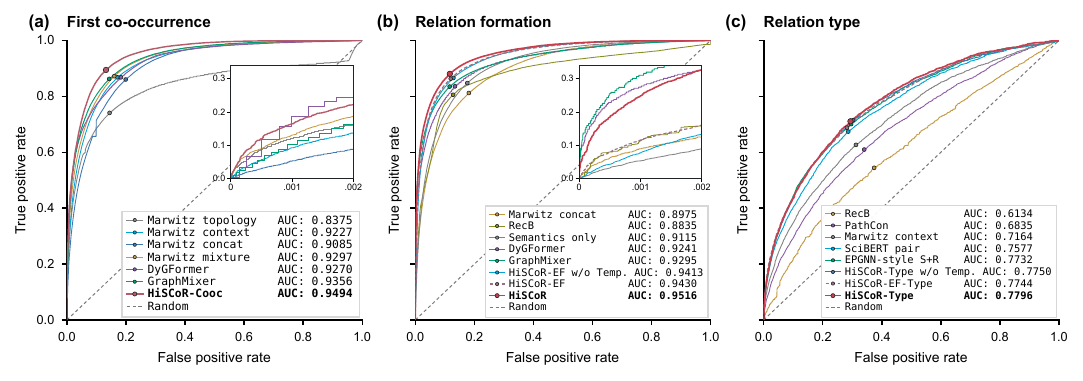}
\caption{\textbf{Held-out test ROC curves.} T0/T1 use population
weights; T2 averages one-vs-rest curves. Insets show low FPR; circles
mark operating points; bold/dashed curves denote final HiSCoR/EF.
Table~\ref{tab:overall_results} reports three-seed means and PR metrics.}
\label{fig:main_v2}
\end{figure*}

\begin{table}[!t]
\centering
\small
\setlength{\tabcolsep}{2pt}
\begin{tabular}{@{}lrrrr@{}}
\specialrule{0.8pt}{0pt}{3pt}
\multicolumn{5}{@{}l}{\bfseries T0\quad First co-occurrence} \\
\cmidrule(lr){1-5}
Model & AUROC & Pop.\ AUPRC & Brier $\downarrow$ & AP-Lift \\
\midrule
Marwitz mixture & 0.9296 & $2.97{\times}10^{-3}$ & $6.02{\times}10^{-2}$ & 69.4$\times$ \\
DyGFormer & 0.9273 & $3.62{\times}10^{-3}$ & $6.92{\times}10^{-2}$ & 84.5$\times$ \\
GraphMixer & 0.9348 & $2.09{\times}10^{-3}$ & $6.07{\times}10^{-2}$ & 48.8$\times$ \\
\rowcolor{oursrow}
HiSCoR-Cooc & & & & \\
\rowcolor{oursrow}
\hspace*{0.7em}w/o Temp.
& \secondscore{0.9420}
& \secondsciscore{3.90}{-3}
& \secondsciscore{7.00}{-5}
& \secondlift{91.1} \\
\rowcolor{oursrow}
\textbf{HiSCoR-Cooc}
& \bestscore{0.9495}
& \bestsciscore{1.028}{-2}
& \bestsciscore{4.29}{-5}
& \bestlift{240.0} \\
\specialrule{0.7pt}{3pt}{3pt}
\multicolumn{5}{@{}l}{\bfseries T1\quad First scientific-relation formation} \\
\cmidrule(lr){1-5}
Model & AUROC & Pop.\ AUPRC & Brier $\downarrow$ & AP-Lift \\
\midrule
Marwitz concat & 0.8982 & $2.156{\times}10^{-3}$ & $6.68{\times}10^{-2}$ & 1131.5$\times$ \\
DyGFormer & 0.9238 & $2.533{\times}10^{-3}$ & $4.25{\times}10^{-2}$ & 1329.5$\times$ \\
GraphMixer & 0.9290 & \secondsciscore{3.384}{-3} & $3.40{\times}10^{-2}$ & \secondlift{1776.1} \\
\rowcolor{oursrow}
HiSCoR-EF & & & & \\
\rowcolor{oursrow}
\hspace*{0.7em}w/o Temp.
& 0.9409
& $2.490{\times}10^{-4}$
& $1.33{\times}10^{-5}$
& 130.7$\times$ \\
\rowcolor{oursrow}
HiSCoR-EF
& \secondscore{0.9436}
& $1.168{\times}10^{-3}$
& \bestsciscore{3.22}{-6}
& 612.8$\times$ \\
\rowcolor{oursrow}
\textbf{HiSCoR}
& \bestscore{0.9515}
& \bestsciscore{3.859}{-3}
& \secondsciscore{4.87}{-6}
& \bestlift{2025.2} \\
\specialrule{0.7pt}{3pt}{3pt}
\multicolumn{5}{@{}l}{\bfseries T2\quad Relation type at formation} \\
\cmidrule(lr){1-5}
Model & \shortstack{Macro\\AUROC} & Macro-AP & Macro-$F_1$ & Rare-$F_1$ \\
\midrule
Marwitz context & 0.7152 & 0.5078 & 0.4872 & 0.0789 \\
SciBERT pair & 0.7549 & 0.5231 & 0.4915 & 0.0901 \\
PathCon & 0.6832 & 0.4626 & 0.4159 & 0.0323 \\
\rowcolor{oursrow}
HiSCoR-Type & & & & \\
\rowcolor{oursrow}
\hspace*{0.7em}w/o Temp.
& 0.7754
& 0.5425
& \secondscore{0.5034}
& 0.0927 \\
\rowcolor{oursrow}
HiSCoR-EF-Type
& \secondscore{0.7768}
& \bestscore{0.5451}
& 0.5033
& \bestscore{0.1087} \\
\rowcolor{oursrow}
\textbf{HiSCoR-Type}
& \bestscore{0.7795}
& \secondscore{0.5431}
& \bestscore{0.5118}
& \secondscore{0.0975} \\
\bottomrule
\end{tabular}
\caption{\textbf{Held-out test performance.} Three-seed means;
best/runner-up are bold/underlined, and blue rows are HiSCoR.}
\label{tab:overall_results}
\end{table}

\begin{table}[!t]
\centering
\small
\setlength{\tabcolsep}{3pt}
\begin{tabular}{@{}lccc@{}}
\toprule
Model & $K=100$ & $K=1$K & $K=10$K \\
\midrule
Marwitz concat
& \underline{7.00 / 0.015}
& \underline{6.00 / 0.126}
& 2.97 / 0.626 \\
GraphMixer
& 5.67 / 0.012
& 5.23 / 0.110
& \underline{4.09 / 0.863} \\
\rowcolor{oursrow}
HiSCoR-EF
& 0.00 / 0.000
& 0.73 / 0.015
& 0.76 / 0.160 \\
\rowcolor{oursrow}
\textbf{HiSCoR}
& \textbf{12.00 / 0.025}
& \textbf{9.00 / 0.190}
& \textbf{4.85 / 1.022} \\
\bottomrule
\end{tabular}
\caption{\textbf{Future-blind T1 candidate reranking.}
Precision@K / end-to-end Recall@K (\%).}
\label{tab:t1_topk}
\end{table}

\section{Experiments}
\label{sec:exp}

We evaluate HiSCoR along three dimensions. First, we compare it with
task-specific baselines at each stage of the hierarchy
(\S\ref{sec:main_results}). Second, we isolate when pre-cutoff history
helps and when static structure suffices (\S\ref{sec:ablation_results}).
Third, we assess hierarchy consistency, calibration, and rare-event
limitations (\S\ref{sec:hierarchy}). Because first co-occurrence and
first relation formation are extremely rare, we emphasise
population-weighted ranking and calibration metrics rather than AUROC
alone. Code, checkpoints, benchmark splits, and redistributable artifacts will be released publicly upon publication.

\subsection{Experimental Setup}

\paragraph{Data splits and population weighting.}
We follow the SCoR-Bench v2 protocol of \S\ref{sec:scor_bench}. T0/T1
evaluation combines a census of positives with uniformly sampled
negatives under inverse-probability weighting. Their fixed test views
contain 11{,}063{,}284 and 11{,}263{,}033 rows, representing about 24.9B
eligible pairs, with positive rates $4.28\times10^{-5}$ and
$1.90\times10^{-6}$; relation formation is more than an order of
magnitude rarer than co-occurrence. T2 contains all 46{,}466
relation-positive test pairs:
30{,}480 \textsc{uses}, 15{,}688 \textsc{combines}, and 298
\textsc{replaces-or-contradicts}.

\paragraph{Baselines.}
Baselines span four families. Scientific-link methods
include Science4Cast \citep{krenn2023forecasting} and the topology,
context, concatenation, and mixture variants of Marwitz et al.\
\citep{marwitz2026predicting}. Continuous-time temporal-graph models
include GraphMixer \citep{cong2023graphmixer} and DyGFormer
\citep{yu2023dygformer}. Relational and temporal-KG references include
DistMult, ComplEx, RotatE, DaeMon, RecB, and ULTRA
\citep{yang2015distmult,trouillon2016complex,sun2019rotate,dong2023daemon,
gastinger2024recb,galkin2024ultra}; T2 adds a SciBERT pair classifier and
PathCon \citep{beltagy2019scibert,wang2021pathcon}. We preserve native
inputs and inductive biases, adapting only the temporal split and output
head.

\begin{figure}[t]
\centering
\includegraphics[width=\columnwidth]{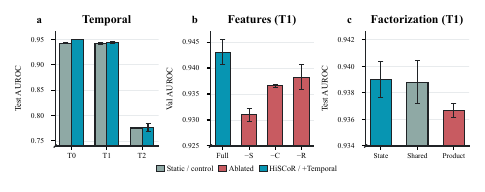}
\caption{\textbf{Matched HiSCoR-family diagnostics.}
(a) Task-specific static/temporal test comparisons (T1: HiSCoR-EF);
(b) validation \textbf{S/C/R} ablations; and (c) HiSCoR-State test
factorization controls. Bars show three-seed means and sample SDs;
these are precursor, not final-model, ablations.}
\label{fig:ablation_overview}
\end{figure}

\paragraph{Implementation and metrics.}
HiSCoR variants train from scratch with AdamW, gradient clipping,
warm-up, and cosine decay (at most 3{,}000 steps; T2: 1{,}600). Runs use
seeds 17/23/42; T2 uses class-balanced batches, while evaluation rows
remain fixed. Checkpoint selection uses validation only: AUROC for T0/T1
diagnostics, population-AUPRC with AUROC $\geq0.94$ for final T1, and
Macro-AP within 0.002 of the best Macro-AUROC for final T2 (all seeds
select step 300). Figure~\ref{fig:ablation_overview}(b) uses its
prespecified Pareto criterion. T0/T1 report AUROC,
population-AUPRC, Brier, and AP-Lift (population-AUPRC divided by the
empirical positive rate); T2 reports Macro-AUROC, Macro-AP,
Macro-$F_1$, and rare-class $F_1$. Unless noted, values are test means
$\pm$ sample SDs over three seeds.

\begin{figure}[t]
\centering
\includegraphics[width=\columnwidth]{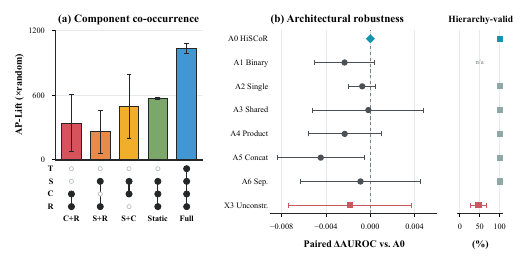}
\caption{\textbf{Matched diagnostic evidence.}
(a) HiSCoR-EF validation AP-Lift for Temporal/\textbf{S/C/R}
(Brier: $1.46\times10^{-5}\!\to3.94\times10^{-6}$).
(b) Frozen-Test paired $\Delta$AUROC and hierarchy validity around
HiSCoR-State A0; X3 violates the hierarchy in 53.06\%. All are
precursor variants.}
\label{fig:mechanistic_evidence}
\end{figure}

\subsection{Overall Performance}
\label{sec:main_results}

Table~\ref{tab:overall_results} and Figure~\ref{fig:main_v2} summarize
performance. Static backbones discriminate broadly; temporal pathways
give strong rare-positive T0/T1 gains but a smaller T2
ranking/rare-class trade-off.

\emph{First co-occurrence (T0).} HiSCoR-Cooc reaches $0.9495\pm0.0005$ AUROC and
$240.0\times$ AP-Lift (Table~\ref{tab:overall_results},
Figure~\ref{fig:main_v2}(a)). Relative to its static counterpart,
population-AUPRC is $2.6\times$ higher and Brier is $39\%$ lower;
both variants outperform GraphMixer.

\emph{Relation formation (T1).} The static HiSCoR backbone already
reaches 0.9409 AUROC but only $2.490\times10^{-4}$ population-AUPRC.
HiSCoR-EF raises it by $4.7\times$; final HiSCoR reaches 0.9515 AUROC
and $3.859\times10^{-3}$ population-AUPRC, exceeding GraphMixer by
0.0225 AUROC and 14.0\% population-AUPRC
(Table~\ref{tab:overall_results}, Figure~\ref{fig:main_v2}(b)).
This gain persists under operational review budgets
(Table~\ref{tab:t1_topk}). Among 74{,}769{,}046 future-blind candidates,
HiSCoR beats GraphMixer for every seed and $K$: mean hits at 100/1K/10K
rise from 5.7/52.3/409.3 to 12.0/90.0/484.7
(112\%/72\%/18\% higher precision), and Recall@10K from 0.863\% to
1.022\%. Against 0.0235\% prevalence, its 12.00\%/9.00\%/4.85\%
precisions yield $511\times$/$383\times$/$206\times$ enrichments;
random ranking yields 0.02/0.23/2.35 hits. Recall uses all 47{,}433 T1
test positives (17{,}561 candidate-retrieved). These results support
rare-positive reranking, although the 37.02\% recall ceiling precludes
a complete open-world discovery claim; all rankings use frozen Test
scores selected on validation.

\emph{Relation type (T2).} Relative to its static counterpart, final
HiSCoR-Type raises Macro-AUROC from 0.7754 to 0.7795 and Macro-$F_1$
from 0.5034 to 0.5118. HiSCoR-EF-Type retains the best Macro-AP
(0.5451) and rare $F_1$ (0.1087), exposing a ranking/rare-class
trade-off; even \textbf{S+R} exceeds the SciBERT pair baseline
(Figure~\ref{fig:main_v2}(c)).

\subsection{Where the Gains Come From}
\label{sec:ablation_results}

Figures~\ref{fig:ablation_overview} and
\ref{fig:mechanistic_evidence} diagnose precursor variants:
HiSCoR-EF supplies static feature/temporal controls, while HiSCoR-State
A0 is compared with matched A1--A6 heads/encoders and unconstrained X3.

\paragraph{Evidence sources are complementary.}
From scratch, HiSCoR-EF reaches $0.9430\pm0.0024$ validation AUROC;
removing \textbf{S}, \textbf{C}, or \textbf{R} lowers it to
$0.9310\pm0.0012$, $0.9366\pm0.0003$, and $0.9382\pm0.0024$
(Figure~\ref{fig:ablation_overview}(b)). Adding temporal evidence
raises AP-Lift from $569.6\pm11.1$ to $1030.3\pm46.8$ and lowers Brier
from $1.46\times10^{-5}$ to $3.94\times10^{-6}$
(Figure~\ref{fig:mechanistic_evidence}(a)). In HiSCoR-State, removing
relation type, direction, or weight costs 0.00147, 0.00128, and 0.00134
test AUROC.

\paragraph{The hierarchy is robust, not universally superior.}
On frozen Test, A0 and its closest alternative A3 differ by only
0.00018 AUROC; their paired 95\% interval crosses zero
($-0.00519$ to $0.00483$; Figure~\ref{fig:mechanistic_evidence}(b)).
This gap is below every field-removal drop, supporting complementarity
and robustness rather than component-wise attribution to final HiSCoR.

\subsection{Hierarchy, Calibration, and Scope}
\label{sec:hierarchy}
\label{sec:scope}

\paragraph{Hierarchy and calibration.}
HiSCoR has zero $P(R)>P(C^\star)$ cases on an unbiased 250k
population-weighted subsample of the 11.3M-row T1 Test view;
hierarchy-aware variants also have zero, while unconstrained
X3 violates the order for $53.06\%\pm21.01\%$ of pairs
(Figure~\ref{fig:mechanistic_evidence}(b)), establishing consistency
rather than ranking superiority.

\paragraph{Rare-event limits.}
At a $1.90\times10^{-6}$ positive rate, T1 remains difficult
despite population-AUPRC and Top-K gains. The 298
\textsc{replaces-or-contradicts} cases remain challenging (rare
$F_1$: $0.1087\pm0.0128$ for HiSCoR-EF-Type, $0.0975\pm0.0178$ for
HiSCoR-Type; Figure~\ref{fig:main_v2}(c)). Candidate generation
recalls 40.41\%/37.02\% of T1 validation/test positives and 19.04\% of
T0 test positives, below the 0.8 gate; the evidence supports
eligible-candidate scoring, not end-to-end discovery.

\section{Conclusion and Limitations}
\label{sec:conclusion}
\label{sec:limitations}

We introduced SCoR, a framework that advances research-direction
discovery from predicting concept co-occurrence to forecasting whether
and how scientific relations will emerge. It unifies the multi-view
SCoR-Graph, leakage-audited SCoR-Bench, and task-adapted HiSCoR family,
which combines static \textbf{S/C/R} structure with pre-cutoff dynamics
under the event hierarchy. Experiments support complementary evidence
sources, temporal gains for rare events, hierarchy-consistent
predictions, and useful reranking under limited review budgets. Current
evidence is limited to \texttt{cs.CV}, one-year horizons, and incomplete
future-blind candidate recall; stronger retrieval, cross-domain
transfer, and rare-relation modeling remain open.

% =====================================================================
\bibliography{aaai2027}

% =====================================================================
% Merged technical supplement.  Keeping it in this same document gives
% arXiv one unambiguous top-level TeX file while preserving the submitted
% supplement title page and appendix layout.
\makeatletter
\addtolength{\aboverulesep}{0.078pt}
\addtolength{\belowrulesep}{0.127pt}
\setlength{\@dblfptop}{0pt plus 1fil}
\setlength{\@dblfpsep}{8pt plus 2fil}
\setlength{\@dblfpbot}{0pt plus 1fil}
\newcommand{\makescorsupplementtitle}{%
  \twocolumn[%
    \vbox to \titlebox {%
      \hsize\textwidth%
      \linewidth\hsize%
      \vskip 0.625in minus 0.125in%
      \centering%
      {\LARGE\bf SCoR: Supplementary Material\par}%
      \vskip 0.1in plus 0.5fil minus 0.05in%
      {\Large{\textbf{%
        \begingroup
          \renewcommand{\thefootnote}{\fnsymbol{footnote}}%
          \hspace*{\fontdimen2\font}Jingze Wang\footnotemark[1],
          Fred Sun\footnotemark[1],
          Shangqi Guo\footnotemark[2]
        \endgroup
        \ifhmode\\\fi
      }}}%
      \vskip .2em plus 0.25fil%
      {\normalsize Center for Brain-Inspired Computing Research,
        Tsinghua University\\}%
      \vskip 1em plus 2fil%
    }%
  ]%
  \begingroup
    \renewcommand{\thefootnote}{\fnsymbol{footnote}}%
    \footnotetext[1]{These authors contributed equally.}%
    \footnotetext[2]{Corresponding author.}%
  \endgroup
  \setcounter{footnote}{0}%
}
\makeatother

\makescorsupplementtitle

\appendix
\setcounter{table}{0}
\setcounter{figure}{6}
\setcounter{equation}{3}
% ===== inlined: sections/method_appendix.tex =====
\section{Technical Details}
\label{app:technical}

\begin{figure*}[!t]
\centering
\includegraphics[width=\textwidth]{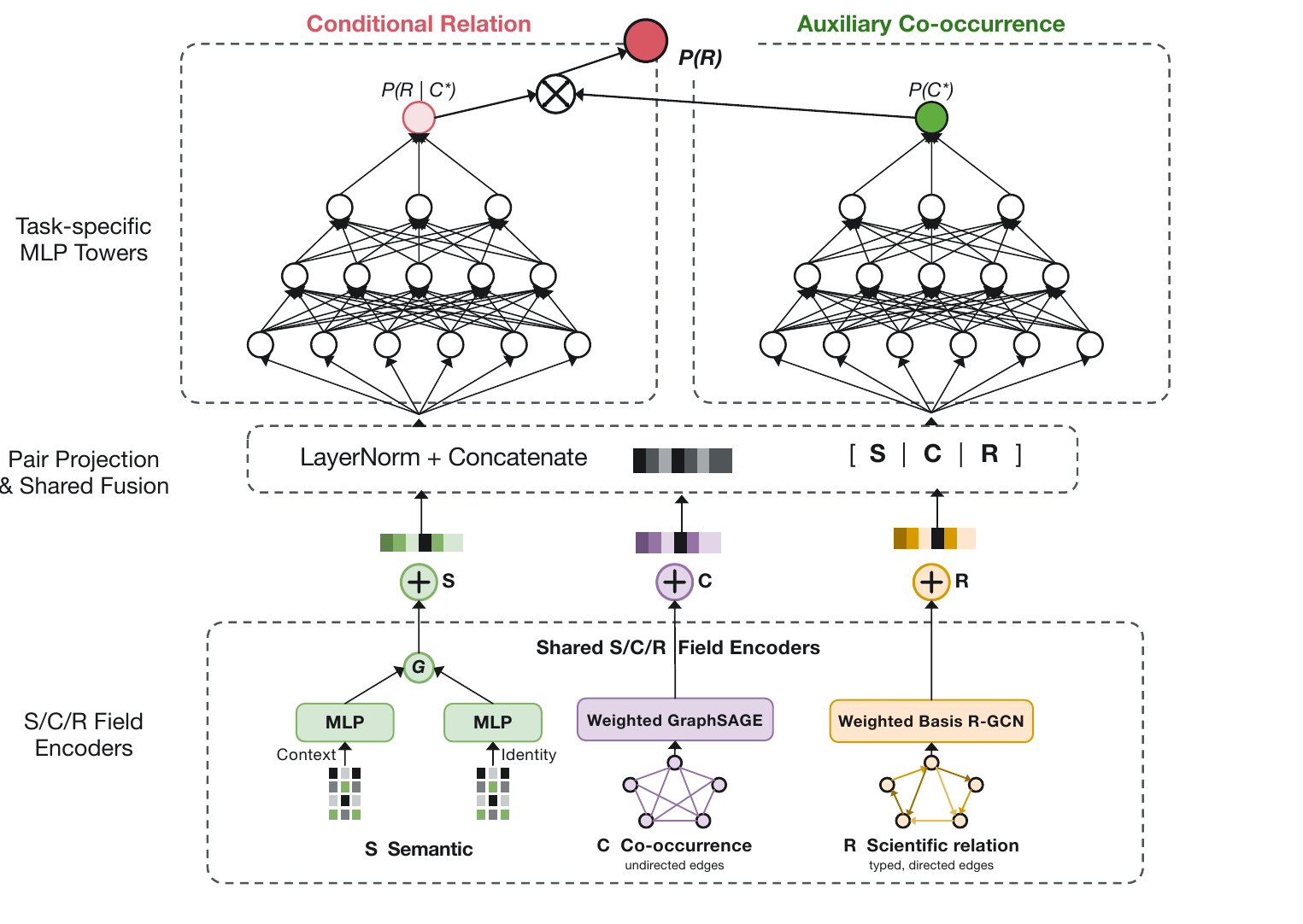}
\caption{\textbf{Original HiSCoR-State architecture.}
Shared semantic (\(\mathcal S\)), co-occurrence (\(\mathcal C\)), and
typed-relation (\(\mathcal R\)) field encoders produce a normalized pair
state for two task-specific MLP towers. The auxiliary tower estimates
target-window co-occurrence \(P(C^\star)\), while the conditional tower
estimates \(P(R\mid C^\star)\). Their product gives
\(P(R)=P(C^\star)P(R\mid C^\star)\), structurally guaranteeing
\(P(R)\le P(C^\star)\). This static hierarchy-consistent model serves as
the backbone for the temporal-event extensions described in the main text.}
\label{fig:hiscor_state_architecture}
\end{figure*}

This appendix records the estimator, encoder, and objective details
omitted from the main text. It does not change the task definitions or
evaluation populations. Figure~\ref{fig:hiscor_state_architecture} presents
the original static HiSCoR-State backbone on which the event-enhanced
variants are built.

\subsection{Population Estimation}
\label{app:population}

For a sampled T0 or T1 row \(i\), let \(\rho_i\) be its inclusion
probability. Population metrics are computed with
\begin{equation}
\begin{aligned}
w_i&=\rho_i^{-1},\\[3pt]
\widehat{\mathcal M}_{\mathrm{pop}}
&=\mathcal M\!\left(
\{(y_i,\hat p_i,w_i)\}_{i=1}^{n}
\right).
\end{aligned}
\label{eq:app_population_weighting}
\end{equation}
All positives in the released evaluation view have their recorded
inclusion probability; sampled negatives retain the probability induced
by the future-blind retrieval and sampling procedure.

\subsection{Hierarchy-Consistent Static State}
\label{app:state}

Let \(\mathbf z_{uv,t}\) be the static pair representation. Two learned
logits induce a distribution over no target-window co-occurrence,
co-occurrence without relation, and relation formation:
\begin{equation}
\begin{aligned}
\boldsymbol{\pi}_{uv,t}
&=\operatorname{softmax}\!\left(
[0,\ell_C(\mathbf z_{uv,t}),\ell_R(\mathbf z_{uv,t})]
\right),\\[2pt]
p^C_{uv,t}&=\pi^C_{uv,t}+\pi^R_{uv,t},\\[2pt]
q^{\mathrm{static}}_{uv,t}
&=\frac{\pi^R_{uv,t}}
{\pi^C_{uv,t}+\pi^R_{uv,t}}.
\end{aligned}
\label{eq:app_static_state}
\end{equation}
The event-corrected conditional probability in
Eq.~(3) of the main paper remains in \([0,1]\), so multiplying it by
\(p^C\) guarantees \(0\le p^R\le p^C\le1\).

\newpage
\subsection{AUPRC-Oriented Temporal Refinement}
\label{app:temporal_refinement}

HiSCoR-EF and HiSCoR-EF-Type are the original conditional and
class-wise event-fusion implementations. Final T1 HiSCoR retains its
static backbone and uses
\begin{equation}
\widetilde g=0.10+0.90g(\mathbf z,\mathbf e),\qquad
\Delta_{\mathrm{event}}
=\operatorname{softplus}(s)\widetilde g\,r(\mathbf z,\mathbf e),
\label{eq:app_temporal_refinement}
\end{equation}
with gate-logit bias \(-0.5\), \(s=0\) at initialization, and Xavier
gain \(0.5\) for the residual output. The event encoder uses 20 recent
tokens and two MLP-Mixer layers. These choices strengthen the temporal
residual without modifying \(p^C\) or the hierarchy.

Final T2 HiSCoR-Type uses a class-wise residual with a separate,
fixed-scale parameterization,
\begin{equation}
\begin{aligned}
\Delta_{\mathrm{event}}^K
&=0.50\,\operatorname{softplus}(\mathbf s_K)\\[-1pt]
&\quad\odot\left[0.10+0.90\sigma\!\left(
g_K(\mathbf z^K,\mathbf e)\right)\right]\odot r_K(\mathbf e),
\end{aligned}
\label{eq:app_t2_temporal_refinement}
\end{equation}
where \(\mathbf s_K=\mathbf 0\) initially and the gate-logit bias is
\(-0.5\). It uses the same 20-token, two-layer event encoder, but the
output of \(r_K\) is initialized from \(\mathcal N(0,10^{-2})\), rather
than with the T1 Xavier initialization.

\subsection{Field Encoders and Pair Construction}
\label{app:fields}

For concept \(v\), separate projections encode its cutoff-safe context
vector \(\mathbf c_v\) and identity vector \(\mathbf i_v\). Their
element-wise gated semantic state is
\begin{equation}
\begin{aligned}
\widetilde{\mathbf c}_v&=f_{\mathrm{ctx}}(\mathbf c_v),\\[2pt]
\widetilde{\mathbf i}_v&=f_{\mathrm{id}}(\mathbf i_v),\\[2pt]
\mathbf g_v&=\sigma\!\left(
W_g[\widetilde{\mathbf c}_v;\widetilde{\mathbf i}_v]+\mathbf b_g
\right),\\[2pt]
\mathbf h_v^{\mathcal S}
&=\operatorname{LN}\!\left(
\mathbf g_v\odot\widetilde{\mathbf c}_v+
(1-\mathbf g_v)\odot\widetilde{\mathbf i}_v
\right).
\end{aligned}
\label{eq:app_semantic_gate}
\end{equation}
A support-weighted GraphSAGE produces
\(\mathbf h_v^{\mathcal C}\), and a weight-aware relational GCN with
forward and inverse channels produces \(\mathbf h_v^{\mathcal R}\).
For each field \(X\in\{\mathcal S,\mathcal C,\mathcal R\}\), endpoint
representations are converted to the symmetric pair state
\begin{equation}
\boldsymbol{\phi}^{X}_{uv}
=\left[
\mathbf h^{X}_{u}+\mathbf h^{X}_{v};
\left|\mathbf h^{X}_{u}-\mathbf h^{X}_{v}\right|;
\mathbf h^{X}_{u}\odot\mathbf h^{X}_{v}
\right].
\label{eq:app_pair_operator}
\end{equation}
The final static state concatenates normalized field projections:
\begin{equation}
\mathbf z_{uv,t}
=\mathop{\Vert}_{X\in\{\mathcal S,\mathcal C,\mathcal R\}}
\operatorname{LN} f_X(\boldsymbol{\phi}^{X}_{uv}).
\label{eq:app_pair_state}
\end{equation}
These \(\mathcal S/\mathcal C/\mathcal R\) fields describe the original
HiSCoR-State model and the final T1 static backbone shown in
Figure~\ref{fig:hiscor_state_architecture}. Final T2 HiSCoR-Type instead
uses an endpoint-conditioned EPGNN relation-type tower: source and target
type vectors, support-weighted neighborhood features, and the
co-occurrence field form its static type logits, with co-occurrence also
serving as an auxiliary expert. Thus, the R-GCN relation encoder in
Figure~\ref{fig:hiscor_state_architecture} is not the final T2 main
relation encoder.

\subsection{Conditional Relation-Type Head}
\label{app:type}

The final T2 type model adds the class-wise event residual in
Eq.~\eqref{eq:app_t2_temporal_refinement} to its static logits:
\begin{equation}
\begin{aligned}
\boldsymbol{\ell}^{K}_{uv,t}
&=a_K(\mathbf z^{K}_{uv,t})\\
&\quad+\Delta_{\mathrm{event}}^K
(\mathbf z^{K}_{uv,t},\mathbf e_{uv,t}),\\[3pt]
\mathbf q^{K}_{uv,t}
&=\operatorname{softmax}(\boldsymbol{\ell}^{K}_{uv,t}).
\end{aligned}
\label{eq:app_type_head}
\end{equation}
Both the gate and residual are class-specific. This lets the same event
history provide different evidence for \textsc{uses},
\textsc{combines}, and \textsc{replaces/contradicts}.

\subsection{Training Objectives}
\label{app:objectives}

For T1, the population-weighted three-state loss is
\begin{equation}
\mathcal L_{\mathrm{state}}
=
\frac{\sum_i w_i\,
\operatorname{CE}(\mathbf s_i,\widetilde{\boldsymbol\pi}_i)}
{\sum_i w_i},
\label{eq:app_state_loss}
\end{equation}
where
\(\widetilde{\boldsymbol\pi}_i=
[1-p_i^C,\ p_i^C-p_i^R,\ p_i^R]\).

The task objectives are
\begin{align}
\mathcal L_{\mathrm{T0}}
&=\mathcal L_{\mathrm{IPW\mbox{-}BCE}}
+\lambda_0\mathcal L_{\mathrm{rank}}^{(0)},\notag\\[3pt]
\mathcal L_{\mathrm{T1}}
&=\mathcal L_{\mathrm{state}}
+\lambda_C\mathcal L_{\mathrm{rank}}^{C}\notag\\[2pt]
&\quad+\lambda_R\mathcal L_{\mathrm{rank}}^{R}
+\lambda_{\mathrm{top}}\mathcal L_{\mathrm{top}}^{R}\notag\\[2pt]
&\quad+\lambda_{\mathrm{temp}}\mathcal L_{\mathrm{rank}}^{\mathrm{temp}}
+\lambda_{\mathrm{temp\mbox{-}top}}
\mathcal L_{\mathrm{top}}^{\mathrm{temp}}\notag\\[2pt]
&\quad+\lambda_{\mathrm{cond}}\mathcal L_{\mathrm{cond}}
+\lambda_{\mathrm{hier}}\mathcal L_{\mathrm{hier}},\notag\\[3pt]
\displaybreak[3]
\mathcal L_{\mathrm{T2}}
&=\operatorname{CE}_{0.03}(\boldsymbol{\ell}^{K},K)
+0.30\operatorname{CE}_{0.03}(\boldsymbol{\ell}^{K}_{\mathrm{sem}},K)\notag\\[2pt]
&\quad+0.20\operatorname{CE}_{0.03}(\boldsymbol{\ell}^{K}_{\mathrm{rel}},K)
+0.20\operatorname{CE}_{0.03}(\boldsymbol{\ell}^{K}_{\mathrm{event}},K)\notag\\[2pt]
&\quad+\lambda_K\mathcal L_{\mathrm{rank}}^K\notag\\[2pt]
&\quad+\lambda_{\mathrm{rare}}
\mathcal L_{\mathrm{rank}}^{\mathrm{rare}}.
\label{eq:app_hiscor_loss}
\end{align}
Here \(w_i\) is the inverse inclusion probability and
\(\operatorname{CE}_{0.03}\) is label-smoothed cross-entropy with
\(\epsilon=0.03\). The ranking losses compare observed positives with
training-only sampled negatives; the top-rank term emphasizes relation
positives near the head of the ranked list. Neither term alters the
validation or test population.

For final T1 HiSCoR, the loss weights are
\[
\begin{aligned}
(&\lambda_C,\lambda_R,\lambda_{\mathrm{top}},
\lambda_{\mathrm{temp}},\\
&\lambda_{\mathrm{temp\mbox{-}top}},\lambda_{\mathrm{cond}},
\lambda_{\mathrm{hier}})
=(0.2,1.0,3.0,0.5,1.0,0.1,1.0).
\end{aligned}
\]
Each batch draws 60\% from the natural population; within the remaining
discriminative stream, relation positives, co-occurrence-only examples,
hard negatives, and random negatives occupy 29/1/50/20\%. Runs start
from random parameters and checkpoints maximize validation
population-AUPRC subject to AUROC \(\ge0.94\); test data are never used
for selection. Under the product parameterization in
Eq.~\eqref{eq:app_static_state}, \(\mathcal L_{\mathrm{hier}}\) is
structurally zero, but is retained and logged in the final implementation.

For final T2 HiSCoR-Type, the loss weights are
\[
(\lambda_K,\lambda_{\mathrm{rare}})=(0.10,0.25).
\]
Three-class balanced batches are used only for training; there are no
additional class weights in the cross-entropy. The T2 ranking terms use
hard-negative fractions \(0.25\) and \(0.10\) (rare class) with margin
\(0.20\). Among checkpoints within 0.002 of the best validation
Macro-AUROC, selection maximizes validation Macro-AP, with rare-class AP
as tie-breaker; Test remains frozen.

\section{Relation Extraction Prompts}
\label{app:relation_prompts}

This section reports the prompt templates used by the relation-extraction
pipeline. Placeholders enclosed in braces are replaced at inference time.
The bilingual gloss in Layer 2 is rendered as the ASCII phrase
\texttt{positive/negative} for pdfLaTeX compatibility; the operational
definitions, schema, and output fields are otherwise unchanged.

\subsection{Entity Extraction Prompt}
\pagebreak
\begin{lstlisting}[style=finalprompt]
You are an expert system in computer vision and machine learning.
Your task is to extract all meaningful research concepts from the paper abstract below.

Rules:
- Respond only as a Python-style list, for example: ['concept 1', 'concept 2', 'concept 3'].
- Include only concepts that are explicitly present in the abstract text.
- Each concept should appear only once.
- Prefer concise noun phrases over full sentences.
- Extract computer-vision concepts such as tasks, methods, model families, architectures, modules, losses, training paradigms, representations, datasets, metrics, modalities, robustness settings, and application domains.
- Be comprehensive: for a normal informative abstract, extract the specific task, method, model or module names, data/model representations, datasets or benchmarks, losses or metrics, modalities, application domains, and important technical phrases when they are explicitly stated.
- Do not return only one to five broad keywords unless the abstract is genuinely very short or contains very few technical concepts.
- Prefer specific phrases like 'semantic image segmentation', 'domain adaptation', 'image translation model', and 'self supervised learning' over broad terms like 'segmentation', 'adaptation', 'model', or 'learning' when the specific phrase appears.
- Normalize plural concepts to singular when natural, for example 'object detectors' -> 'object detector'.
- Remove punctuation unless it is part of a canonical model, dataset, metric, or method name.
- Expand coordinated concepts consistently when the text clearly implies both, for example 'image and video segmentation' -> 'image segmentation' and 'video segmentation'.
- Keep canonical names such as 'CLIP', 'DINOv2', 'YOLO', 'Faster R-CNN', 'U-Net', 'ImageNet', 'COCO', 'IoU', and 'mAP' when they appear.
- Do not include standalone numbers, years, percentages, distances, metric values, speedups, list indices, equation fragments, or measurement fragments as concepts.
- Do not include standalone '2D' or '3D'; keep them only when they are part of a specific phrase such as '3D object detection', '2D pose estimation', or '3D LiDAR camera calibration'.
- Do not split mathematical attack names into numeric fragments. For example, extract 'l2 adversarial attack' or 'linfinity adversarial attack' if explicitly present, but never extract 'l 2', '1', '2', '3', 'at least 1.8', or 'up to 3'.
- Do not include author names, affiliation names, paper venue names, or generic filler phrases.
- Do not explain your answer.

Abstract:
{ABSTRACT}}
\end{lstlisting}

\subsection{Relation Generator Prompt}
\begin{lstlisting}[style=finalprompt]
You are a scientific relation classifier. You are given TWO technical concepts that co-occur in ONE sentence of a paper, plus that sentence. Decide the relation as a NESTED 3-layer JSON triple, advancing layer by layer.

LAYER 1 - related (0/1): Co-occurrence in a sentence is NOT a relation. Output 1 ONLY if the sentence explicitly states a relation between the two concepts. If they are merely listed, compared in passing, or unrelated, output 0 and stop (set the deeper layers to "none").

LAYER 2 - polarity.value (positive/negative): the positive/negative orientation. NOT sentiment - it is structural:
- positive = CONSTRUCTIVE: the two concepts are built together / one is used inside the other (these grow the field).
- negative = SELECTIVE: one concept supersedes, replaces, beats, or refutes the other (competitive / selection pressure).

LAYER 3 - polarity.relation.type (choose exactly one) and its head->tail direction:
- uses (POSITIVE/constructive): head adopts/employs/builds on tail as a component (head=user, tail=component)
- combines (POSITIVE/constructive) [symmetric]: head and tail are integrated together as co-equal components (symmetric)
- replaces (NEGATIVE/selective): head is the NEWER thing that supersedes tail as a drop-in (head=newer, tail=older)
- contradicts (NEGATIVE/selective): head shows a limitation/failure of or refutes tail (head=critic, tail=criticized)
  direction: for asymmetric relations use "e_i->e_j" (e_i is head/subject) or "e_j->e_i"; for symmetric relations (combines) use "symmetric".

The output is a NESTED triple - each layer is entered only if the previous one passes (do NOT flatten the layers):

{
  "e_i": "<id>", "e_j": "<id>",
  "related": 0 or 1,
  "polarity": {
    "value": "positive" | "negative" | "none",
    "relation": {
      "type": "<relation>" | "none",
      "direction": "e_i->e_j" | "e_j->e_i" | "symmetric" | "none"
    }
  },
  "confidence": 0..1,
  "evidence_span": "<verbatim substring of the sentence>",
  "rationale": "<one sentence>"
}

RULES:
1. evidence_span MUST be a verbatim substring of the provided sentence. Never invent text.
2. A relation whose tail is a dataset, benchmark, metric, training infrastructure, or raw input/output data or data modality is NOT a concept-concept relation -> related=0.
3. `uses` requires the head to ACTIVELY EMPLOY tail as a COMPONENT/method in this work. Data inputs, prediction/optimization targets, historical configurations, and mere co-occurrence are not `uses`.
4. Only assert what is EXPLICITLY stated; do not infer beyond the sentence.
5. Components integrated into ONE method/model/framework/pipeline are `combines` (symmetric); co-listed datasets, benchmarks, tasks, metrics, or compared methods are not.
6. confidence in [0,1] = your calibrated probability the full triple is correct.

Output ONLY the JSON object - no prose, no markdown fences.
\end{lstlisting}

\subsection{Relation Discriminator Prompt}
\begin{lstlisting}[style=finalprompt]
Classify the relation between e_i and e_j based on the sentence.
{
  "e_i": "<e_i>",
  "e_i_surface": "<concept_i>",
  "e_j": "<e_j>",
  "e_j_surface": "<concept_j>",
  "sentence": "<sentence>",
  "comparative_cues_present": [],
  "candidate_kind": "cooccur"
}

Return the nested 3-layer JSON triple. e_i and e_j in your output must match the ids above exactly.
\end{lstlisting}

\section{Final Data and Reproducibility Record}
\label{app:final_record}
\renewcommand{\thetable}{S\arabic{table}}

This section records only the final data assets, validation-selected
checkpoints, and frozen test results used by the submitted paper. It
excludes precursor diagnostics, development runs, and validation metrics
that were not used in the final tables.

\begin{table*}[!t]
\centering
\scriptsize
\setlength{\tabcolsep}{4pt}
\renewcommand{\arraystretch}{1.06}
\caption{\textbf{Final SCoR-Graph construction and identity-model record.}
U/C/R/X denotes \textsc{uses}/\textsc{combines}/\textsc{replaces}/
\textsc{contradicts}. Identity rows report separately held-out test
performance; the parentheses in the final two rows give selected-threshold
precision/recall.}
\label{tab:app_graph_record}
\begin{tabular}{@{}lr@{\hspace{1.4em}}lr@{}}
\toprule
Statistic & Value & Statistic & Value\\
\midrule
Papers & 187,848 & Coverage & 2017-01-03 to 2026-06-25\\
Surface concepts & 290,197 & Active surface concepts & 290,001\\
Merged surface concepts & 19,510 & Final graph nodes & 270,687\\
Blocked ambiguous acronym families & 975 & Unique co-occurrence edges & 7,452,716\\
Co-occurrence paper events & 9,199,848 & Unique typed relation edges & 615,036\\
Underlying relation pairs & 580,615 & Relation-evidence events & 1,306,721\\
Relation paper edges & 660,489 & Typed edges (U/C/R/X) &
440,689 / 168,982 / 4,487 / 878\\
\midrule
V5 bi-encoder (n; AUROC; AP) & 4,663; 0.9934; 0.9858 &
V6 verifier (n; AUROC; AP) & 6,760; 0.9959; 0.9854\\
V5 threshold (precision/recall) & 0.9758 (0.991/0.755) &
V6 threshold (precision/recall) & 0.9990 (0.994/0.843)\\
\bottomrule
\end{tabular}
\end{table*}

\begin{table*}[!t]
\centering
\scriptsize
\setlength{\tabcolsep}{4pt}
\renewcommand{\arraystretch}{1.08}
\caption{\textbf{Final SCoR-Bench chronological splits and frozen test
views.} Master rows are the final benchmark rows before task-specific
evaluation views. U/C/R denotes \textsc{uses}/\textsc{combines}/rare
(\textsc{replaces} or \textsc{contradicts}) T2 positives.}
\label{tab:app_benchmark_record}
\begin{tabular}[t]{@{}lccr@{}}
\toprule
Split & Cutoff & Horizon end & Master rows\\
\midrule
Train-1 & 2022-06-25 & 2023-06-25 & 42,121,503\\
Train-2 & 2023-06-25 & 2024-06-25 & 51,506,057\\
Validation & 2024-06-25 & 2025-06-25 & 63,476,038\\
Test & 2025-06-25 & 2026-06-25 & 75,632,357\\
\bottomrule
\end{tabular}
\hspace{2.2em}
\begin{tabular}[t]{@{}lrrrr@{}}
\toprule
Task & View rows & Eligible pairs & Positives & Positive rate\\
\midrule
T0 & 11,063,284 & 24.899B & 1,065,821 & $4.28{\times}10^{-5}$\\
T1 & 11,263,033 & $\sim$24.9B & 47,433 & $1.90{\times}10^{-6}$\\
T2 & 46,466 & all relation-positive pairs &
30,480 / 15,688 / 298 & ---\\
\bottomrule
\end{tabular}
\end{table*}

\begin{table*}[!t]
\centering
\scriptsize
\setlength{\tabcolsep}{3.5pt}
\renewcommand{\arraystretch}{1.08}
\caption{\textbf{Final HiSCoR-family frozen-test summary (mean $\pm$ sample
SD over seeds 17, 23, and 42).} For T0/T1, metric columns are AUROC,
population AUPRC, Brier, and AP-Lift. For T2, they are Macro-AUROC,
Macro-AP, Macro-F1, and rare-class F1, respectively. Checkpoints were
selected only on validation data.}
\label{tab:app_final_family}
\begin{tabular}{@{}llcccc@{}}
\toprule
Task & Model & AUROC / M-AUROC & AUPRC / M-AP & Brier / M-F1 & AP-Lift / rare-F1\\
\midrule
T0 & HiSCoR-Cooc w/o Temp. &
$0.9420{\pm}0.0012$ & $0.003901{\pm}0.000898$ &
$(7.0034{\pm}0.7520){\times}10^{-5}$ & $91.1{\pm}21.0$\\
T0 & HiSCoR-Cooc &
$0.9495{\pm}0.0005$ & $0.010275{\pm}0.001495$ &
$(4.2937{\pm}0.0106){\times}10^{-5}$ & 240.0\\
\midrule
T1 & HiSCoR-EF w/o Temp. &
$0.9409{\pm}0.0020$ & $0.000249{\pm}0.000152$ &
$(1.3296{\pm}0.0387){\times}10^{-5}$ & $130.7{\pm}79.5$\\
T1 & HiSCoR-EF &
$0.9436{\pm}0.0022$ & $0.001168{\pm}0.000326$ &
$(3.2213{\pm}0.0748){\times}10^{-6}$ & $612.8{\pm}170.9$\\
T1 & HiSCoR &
$0.9515{\pm}0.0001$ & $0.003859{\pm}0.000673$ &
$(4.8725{\pm}0.3562){\times}10^{-6}$ & $2025.2{\pm}353.1$\\
\midrule
T2 & HiSCoR-Type w/o Temp. &
$0.7754{\pm}0.0011$ & $0.5425{\pm}0.0032$ &
$0.5034{\pm}0.0083$ & $0.0927{\pm}0.0017$\\
T2 & HiSCoR-EF-Type &
$0.7768{\pm}0.0078$ & $0.5451{\pm}0.0005$ &
$0.5033{\pm}0.0067$ & $0.1087{\pm}0.0128$\\
T2 & HiSCoR-Type &
$0.7795{\pm}0.0019$ & $0.5431{\pm}0.0026$ &
$0.5118{\pm}0.0081$ & $0.0975{\pm}0.0178$\\
\bottomrule
\end{tabular}
\end{table*}

\begin{table*}[!t]
\centering
\small
\setlength{\tabcolsep}{3.5pt}
\renewcommand{\arraystretch}{1.08}
\caption{\textbf{Validation-selected checkpoints and future-blind T1 Top-$K$
audit.} Panel (a) gives the selected checkpoint steps for seeds 17, 23, and
42. Panel (b) reports means $\pm$ sample SD across the same seeds. The
74,769,046-pair candidate universe retrieves 17,561 of 47,433 test positives
(37.02\% candidate recall); end-to-end recall uses all 47,433 positives.}
\label{tab:app_topk}
\begin{minipage}[t]{0.34\textwidth}
\centering
\textbf{(a) Selected checkpoint step}\\[3pt]
\begin{tabular}{@{}lrrr@{}}
\toprule
Task & Seed 17 & Seed 23 & Seed 42\\
\midrule
T0 HiSCoR-Cooc & 800 & 600 & 2,600\\
T1 HiSCoR & 2,400 & 2,200 & 2,400\\
T2 HiSCoR-Type & 300 & 300 & 300\\
\bottomrule
\end{tabular}
\end{minipage}\hfill
\begin{minipage}[t]{0.63\textwidth}
\centering
\textbf{(b) Future-blind T1 Top-$K$ audit}\\[3pt]
\begin{tabular}{@{}llrrr@{}}
\toprule
Model & $K$ & Hits & Precision & End-to-end recall\\
\midrule
GraphMixer & 100 & $5.7{\pm}0.6$ & $5.67{\pm}0.58$\% & $0.012{\pm}0.001$\%\\
GraphMixer & 1,000 & $52.3{\pm}10.4$ & $5.23{\pm}1.04$\% & $0.110{\pm}0.022$\%\\
GraphMixer & 10,000 & $409.3{\pm}11.5$ & $4.09{\pm}0.12$\% & $0.863{\pm}0.024$\%\\
\midrule
HiSCoR & 100 & $12.0{\pm}4.4$ & $12.00{\pm}4.36$\% & $0.025{\pm}0.009$\%\\
HiSCoR & 1,000 & $90.0{\pm}10.4$ & $9.00{\pm}1.04$\% & $0.190{\pm}0.022$\%\\
HiSCoR & 10,000 & $484.7{\pm}47.6$ & $4.85{\pm}0.48$\% & $1.022{\pm}0.100$\%\\
\bottomrule
\end{tabular}
\end{minipage}
\end{table*}

\end{document}